\pdfoutput=1

\documentclass[11pt]{article}

\usepackage[]{acl}

\usepackage{times}
\usepackage{latexsym}

\usepackage[T1]{fontenc}

\usepackage[utf8]{inputenc}

\usepackage{microtype}

\usepackage{inconsolata}

\usepackage{booktabs}
\usepackage{graphicx}
\usepackage{multirow}
\usepackage{amsmath}
\usepackage{amssymb}
\usepackage{marvosym}
\usepackage{footmisc}
\usepackage{algorithm}
\usepackage{algpseudocode}
\usepackage[breakable]{tcolorbox}

\usepackage{pifont}
\newcommand{\cmark}{\ding{51}} 
\newcommand{\xmark}{\ding{55}} 

\usepackage{color,colortbl}
\definecolor{color2}{rgb}{0,0.5,0}

\newcommand{\ours}{DAMC}
\newcommand{\baseline}{NaiveMC}
\newcommand{\benchmark}{MCUB}

\title{Model Composition for Multimodal Large Language Models}

\author{
Chi Chen\textsuperscript{*,1}, 
Yiyang Du\textsuperscript{*,1}, 
Zheng Fang\textsuperscript{1}, 
Ziyue Wang\textsuperscript{1},
Fuwen Luo\textsuperscript{1}, \\ 
{\bf
Peng Li \textsuperscript{\Letter,2,4}, 
Ming Yan\textsuperscript{3}, 
Ji Zhang\textsuperscript{3}, 
Fei Huang\textsuperscript{3}, 
Maosong Sun\textsuperscript{1}, 
Yang Liu\textsuperscript{\Letter,1,2,4,5}} \\
  \textsuperscript{1}Dept. of Comp. Sci. \& Tech., Institute for AI, Tsinghua University, Beijing, China \\
  \textsuperscript{2}Institute for AI Industry Research (AIR), Tsinghua University, Beijing, China \\
  \textsuperscript{3}Institute of Intelligent Computing, Alibaba Group\\
  \textsuperscript{4}Shanghai Artificial Intelligence Laboratory, Shanghai, China\\
  \textsuperscript{5}Jiangsu Collaborative Innovation Center for Language Competence, Jiangsu, China
}

\begin{document}
\maketitle

\DefineFNsymbols*{1}{*}
\setfnsymbol{1}

\renewcommand{\thefootnote}{\fnsymbol{footnote}} 
    \footnotetext[1]{Equal contribution.}
\renewcommand{\thefootnote}{\arabic{footnote}}

\DefineFNsymbols*{1}{\Letter}
\setfnsymbol{1}

\renewcommand{\thefootnote}{\fnsymbol{footnote}} 
    \footnotetext[1]{Corresponding authors: Peng Li (lipeng@air.tsinghua.\\edu.cn) and Yang Liu (liuyang2011@tsinghua.edu.cn).}
\renewcommand{\thefootnote}{\arabic{footnote}}

\begin{abstract}
Recent developments in Multimodal Large Language Models (MLLMs) have shown rapid progress, moving towards the goal of creating versatile MLLMs that understand inputs from various modalities. However, existing methods typically rely on joint training with paired multimodal instruction data, which is resource-intensive and challenging to extend to new modalities. In this paper, we propose a new paradigm through the model composition of existing MLLMs to create a new model that retains the modal understanding capabilities of each original model. Our basic implementation, \baseline{}, demonstrates the effectiveness of this paradigm by reusing modality encoders and merging LLM parameters. Furthermore, we introduce \ours{} to address parameter interference and mismatch issues during the merging process, thereby enhancing the model performance. To facilitate research in this area, we propose \benchmark{}, a benchmark for assessing ability of MLLMs to understand inputs from diverse modalities. Experiments on this benchmark and four other multimodal understanding tasks show significant improvements over baselines, proving that model composition can create a versatile model capable of processing inputs from multiple modalities.\footnote{Code is available at \url{https://github.com/THUNLP-MT/ModelCompose}}
\end{abstract}

\section{Introduction}

\begin{figure}[t]
    \centering
    \includegraphics[width=1.0\linewidth]{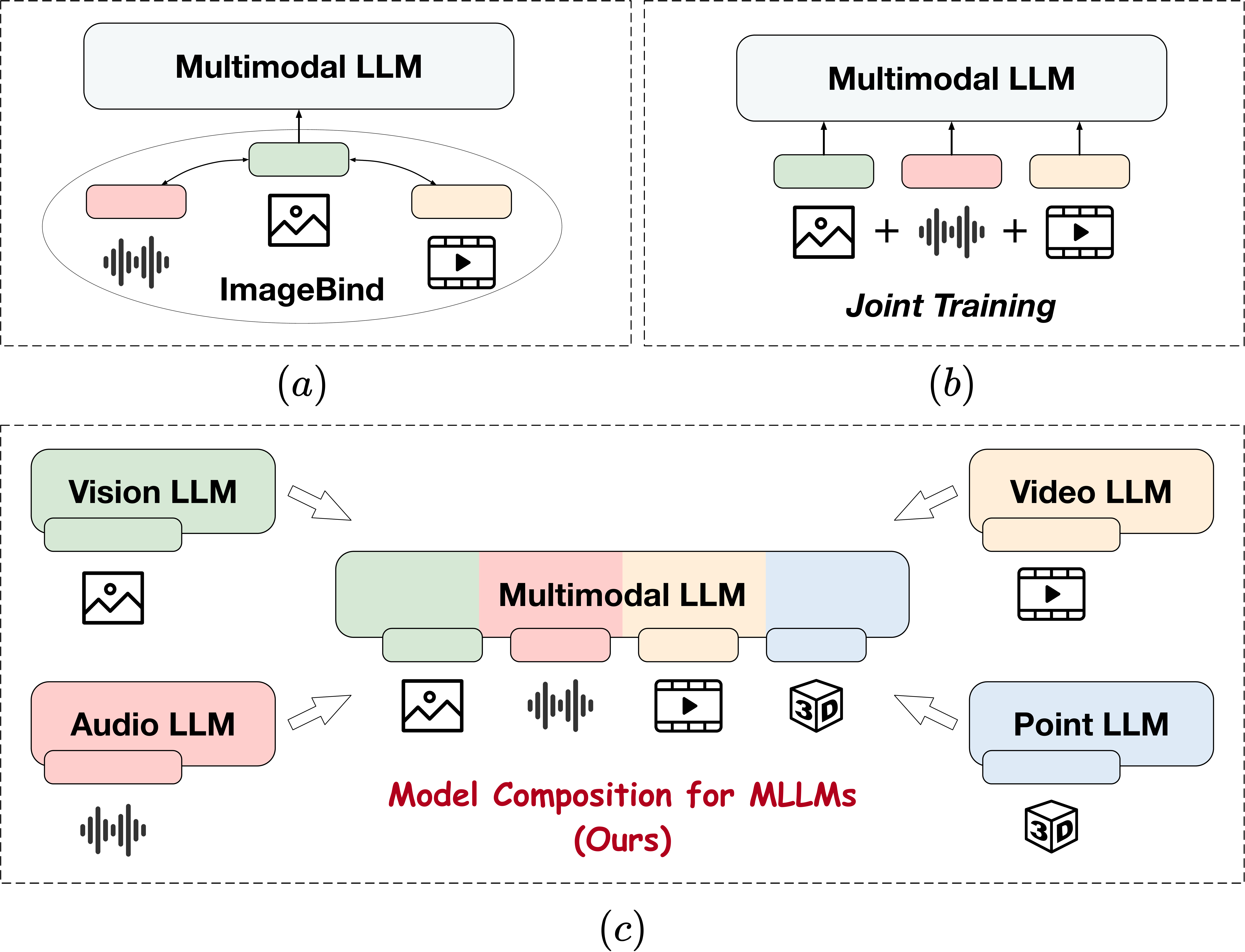}
    \caption{Illustration of various approaches for multimodal large language models: (a) aligning LLM with a multimodal encoder and (b) joint training with multiple modal encoders and (c) our proposed model composition method that creates a versatile model from existing MLLMs through a training-free and extensible process.}
    \label{fig:figure1}
\end{figure}

Recent advancements in Multimodal Large Language Models~(MLLMs) have established them as the forefront of multimodal learning paradigms~\cite{liu2023visual,liu2023improved}. The prevalent approach involves aligning modality encoders with large language models~(LLMs) through extensive modality-text paired data and then fine-tuning with modality-specific instruction data. This paradigm has been successfully applied to a wide range of modalities such as image~\cite{liu2023visual,dai2023instructblip}, audio~\cite{gong2023ltu,deshmukh2023pengi,tang2024salmonn}, video~\cite{zhang2023videollama,lin2023videollava}, and point cloud~\cite{hong20233dllm,xu2023pointllm}, resulting in the emergence of a diverse array of MLLMs with unique modal capabilities. 

There are also some efforts to enable a single MLLM to handle multiple modalities. One way to achieve this is to align the LLM with a multimodal encoder such as ImageBind~\cite{girdhar2023imagebind} with only image-text data~(Figure \ref{fig:figure1}(a))~\cite{su2023pandagpt, han2023imagebindllm}. This approach leverages the inherent alignment of different modalities within the multimodal encoder, allowing the MLLM to comprehend various modalities to a certain degree. However, the absence of modality-specific instruction data often results in suboptimal performance. Another approach entails the concurrent training of the MLLM with multiple modality encoders~(Figure \ref{fig:figure1}(b)). For example, ChatBridge~\cite{zhao2023chatbridge} connects image, video and audio encoders with the LLM through a joint training process with multimodal instruction data~(i.e., video-audio chats). 
This kind of methods show potential but faces two major challenges. First, it is resource-heavy to collect paired data across multiple modalities. Second, adapting these models to new modalities requires additional training, adding to the complexity and resource demands of the development process.

Given the limitations of current approaches, we propose and study a more practical setting: \textit{model composition} for MLLMs~(Figure \ref{fig:figure1}(c)). Our primary research question is simple yet fundamental: \textit{Can we create a new MLLM by combining existing MLLMs to inherit their understanding of different modalities without training?} Model composition for MLLMs is advantageous for two key reasons: (1) it eliminates the need for the resource-heavy process of training and gathering multimodal data, and (2) it promises enhanced adaptability, facilitating seamless incorporation of new modalities.

Some recent studies, such as X-InstructBLIP \cite{panagopoulou2023xinstruct}, serve as pioneering efforts in model composition for MLLMs. These works primarily train projectors to align different encoders with a single LLM and demonstrate the ability to process multiple modalities concurrently. However, a critical limitation is their applicability only to MLLMs with \textbf{frozen} language model weights. This constraint restricts the range of models that can be utilized, and impairs the overall performance of the MLLMs~\cite{zeng2023lynx}.

In this paper, we first propose a framework for model composition for MLLMs. Our implementation, named \baseline{}, is elegantly simple: for the MLLMs to be composed, we directly reuse their modality-specific encoders and merge their LLM parameters. We demonstrate that MLLMs, as long as initialized from the same LLM, can achieve zero-shot multi-modality expansion through this model composition framework, regardless of whether the parameters of the LLM have been fine-tuned. 

Furthermore, to mitigate parameter interference in the composition process and optimize the performance of the composite model, we propose \ours{}, an advanced framework with parameter \textbf{D}ecoupling and \textbf{A}djustment for \textbf{M}odel \textbf{C}omposition. By separating modality-specific parameters from language model parameters during initial MLLM training, \ours{} allows for the selective merging of textual parameters, reducing cross-modal interference. Moreover, \ours{} introduces an adaptive parameter adjustment mechanism to ensure optimal compatibility and effectiveness of the composite model, achieving a balanced and efficient multi-modality expansion.

To assess the efficacy of our proposed frameworks, we conduct comprehensive experiments on tasks that require an integrated understanding of inputs from diverse combinations of four prevalent modalities: image, audio, video, and point cloud. To facilitate the research in model composition for MLLMs, we also build \benchmark{}, a benchmark specifically designed for evaluating the capability to concurrently comprehend multiple modalities by identifying commonalities across inputs from various modalities. Experimental results indicate that our frameworks enable the composition of existing MLLMs from different modalities without requiring further training, yielding a versatile and high-performing multimodal model adept at handling any combination of these modalities.

Our contributions are three-fold:
\begin{itemize}
\setlength{\itemsep}{0pt}
\setlength{\parsep}{0pt}
\setlength{\parskip}{0pt}
    \item We propose the concept of model composition for MLLMs, realized through the \baseline{} framework, which allows for seamless integration of different MLLMs without additional training, enabling zero-shot multi-modality expansion.
    \item We introduce \ours{}, an advanced model composition framework that employs parameter decoupling and adaptive adjustment to mitigate parameter interference and optimize composite model performance across multiple modalities.
    \item We create \benchmark{}, a benchmark designed to evaluate the unified understanding of diverse modalities, and demonstrate the efficacy of our model composition frameworks via extensive experiments on various multimodal understanding tasks and \benchmark{}.
\end{itemize}

\section{Related Work}
\subsection{Multimodal Large Language Models}

Recent advancements have integrated Large Language Models (LLMs) with multiple modalities such as image, video, audio, and point cloud. Approaches like X-LLM \cite{chen2023xllm} utilize modality-specific adapters, while ChatBridge \cite{zhao2023chatbridge} employs a Perceiver for each modality. Macaw-LLM \cite{lyu2023macaw} uses a unified alignment module, and NeXT-GPT \cite{wu2023nextgpt} relies on linear projection for multimodal integration. However, these methods often necessitate joint multimodal dataset training, posing scalability and extension challenges. While PandaGPT \cite{su2023pandagpt} and ImageBind-LLM \cite{han2023imagebindllm} bypass joint training with ImageBind~\cite{girdhar2023imagebind} as a unified encoder, their performance is limited by a lack of modality-specific instruction data. In contrast, our approach benefits from instruction-following training for each MLLM and integrates their capabilities through a training-free process.

\subsection{Model Composition}

Model composition integrates different models' capabilities, primarily through weight space merging strategies~\cite{ilharco2022taskarith,matena2022merging,jin2023dataless,huang2023lorahub,yu2024language,wortsman2022robust}, enhancing fine-tuned models from the same initialization. However, existing research has not fully explored model composition for  multimodal LLMs. Prior works on multimodal tasks use linear interpolation for tasks like speech recognition~\cite{sundar2023multimodal}. Another work converts model processing images and text separately into one using same parameters for simultaneous processing~\cite{sung2023empirical}. But they do not expand the processing modalities of the model. X-InstructBlip \cite{panagopoulou2023xinstruct} incorporates new modalities by training separate projections for modal encoders and LLM, but its performance is limited by the frozen LLM. Our work advances multimodal capabilities by enabling LLM training and applying model composition techniques to LLM weights, significantly improving the effect of multimodal integration.

\begin{figure*}[t]
    \centering
    \includegraphics[width=0.95\textwidth]{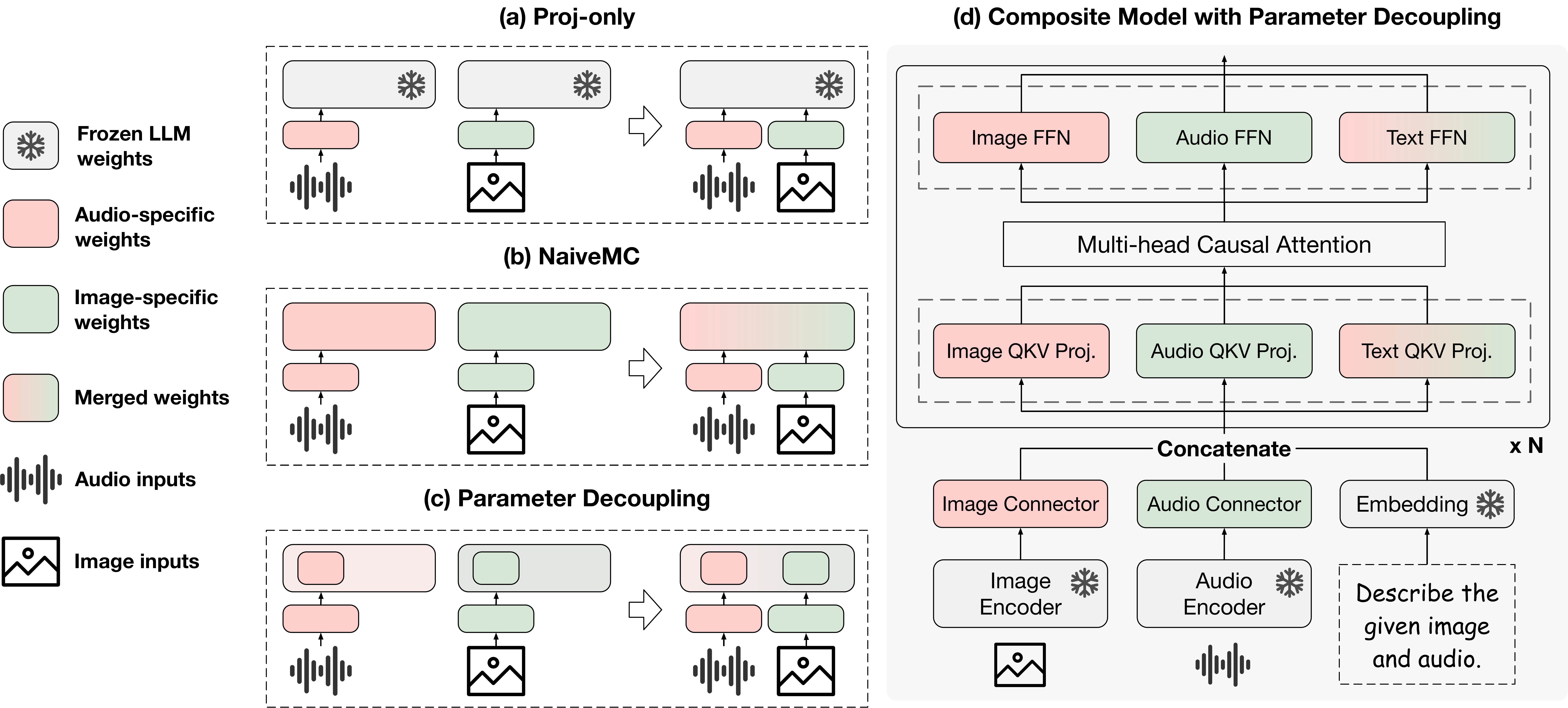}
    \caption{Illustration of the model composition processes with only image and audio modalities are considered for simplicity. (a) and (b) show a basic model composition framework as described in Section \ref{sec:naivemc}, while (c) and (d) demonstrate model composition with parameter decoupling, as detailed in Section \ref{sec:pd}.}
    \label{fig:arch}
\end{figure*}

\section{Methodology}

\subsection{Task Definition}

In this work, we examine the composition of a set of Multimodal Large Language Models (MLLMs), denoted as $\{ M_1, M_2, \ldots, M_n \}$, each capable of responding to textual queries with different modalities $\mathbf{m} = \{ m_1, m_2, \ldots, m_n \}$. The core objective is to develop a composition method $C$ that effectively integrates these MLLMs into a singular, more versatile model $M_\text{compose} = C({ M_1, M_2, \ldots, M_n })$. This integration aims to enable the model to process and understand inputs from any combination of modalities in $\mathbf{m}$. For example, by integrating two specialized MLLMs—a vision LLM and an audio LLM—the resulting composite model should not only preserve the individual proficiencies of these models in processing images and audio, respectively, but should also acquire a zero-shot capacity for handling inputs that encompass both visual and auditory information simultaneously.

\begin{algorithm}
\caption{Model Composition for MLLMs}
\label{alg:composition}
\begin{algorithmic}[1]
\Require Set of MLLMs $\{ M_1, M_2, \ldots, M_n \}$, each with a set of parameters $\Theta^i$
\Ensure Integrated parameters $\Theta_{\text{compose}}$

\State Initialize $\Theta_{\text{common}}$ and $\Theta_{\text{compose}}$ as empty sets
\State Define a mapping $f: \theta \mapsto G$ that maps the parameter $\theta$ to group $G$ based on functionality
\For{$i = 1$ to $n$}
    \For{each parameter $\theta$ in $\Theta^i$}
        \If{$\forall j \neq i, \theta \notin \Theta^j$}
            \State Add $\theta$ to $\Theta_{\text{compose}}$
        \Else
            \State Assign $\theta$ to group $f(\theta)$ in $\Theta_{\text{common}}$
        \EndIf
    \EndFor
\EndFor
\For{each group $G$ in $\Theta_{\text{common}}$}
    \State $\theta_{\text{merge}} = \text{average}(\text{parameters in } G)$
    \State Add $\theta_{\text{merge}}$ to $\Theta_{\text{compose}}$
\EndFor
\State \Return $\Theta_{\text{compose}}$
\end{algorithmic}
\end{algorithm}

\subsection{A Model Composition Framework}
\label{sec:naivemc}

When composing models, two critical elements must be taken into account: the \textbf{components} and their \textbf{weights}.
The prevalent MLLMs typically feature two component types: (1) \textit{modal-specific} components, like modal encoders and projectors, which adapt modality inputs to the language embedding space and (2) \textit{modal-agnostic} components that exist in each MLLM, primarily the underlying LLM itself. In our composition framework, we retain all modal-specific components~(and their weights) from different MLLMs to handle respective modal inputs, and connect them to the same LLM. In cases where the LLMs have not been adapted during the training of MLLMs (as illustrated in Figure \ref{fig:arch}(a)), we employ the pre-trained weights of the LLM directly. Conversely, if the LLMs have undergone adaptation in the MLLM training process~(Figure \ref{fig:arch}(b)), we simply average their weights. We name this composition framework \baseline{} and provide a formal procedure of it in Algorithm \ref{alg:composition}.

\subsection{Parameter Decoupling}
\label{sec:pd}

In the previously described framework, there arises an unaddressed issue, specifically the potential for parameter interference when merging fine-tuned LLM parameters. During the training of MLLMs, the LLM parameters are optimized for specific modal inputs. This specialized optimization could lead to variations in the LLM parameters of different MLLMs. When these parameters are merged, conflicts may arise, potentially impacting the ability of the model to understand modal inputs.

To alleviate the issue of parameter interference during model composition, we advocate for initially training the MLLMs with a parameter decoupling strategy in the first place. As shown in Figure \ref{fig:arch}(c) and Figure \ref{fig:arch}(d), the main idea is to separate the modality processing parameters from those of the language model within MLLMs. For example, in an MLLM $M$ that processes modality $m$, the input for each attention layer is denoted as $X = [X_m, X_t]$ where $X_m$ and $X_t$ represent the modality-specific and text sequences, respectively. The attention components are computed as follows:
\begin{align}
Q = [X_{m}W_{m}^{Q}, X_{t}W_{t}^{Q}] \\
K = [X_{m}W_{m}^{K}, X_{t}W_{t}^{K}] \\
V = [X_{m}W_{m}^{V}, X_{t}W_{t}^{V}]
\end{align}
where $Q$, $K$, and $V$ represent the queries, keys, and values in the attention mechanism, respectively. Note that two sets of weights $W_m^{(\cdot)}$ and $W_t^{(\cdot)}$ are employed, tailored to the modality-specific and textual inputs. The attention operation is then applied:
\begin{equation}
[X_m^{O}, X_{t}^O] = \text{split}(\text{Attention}(Q, K, V))
\end{equation}
which splits the attention output into modality-specific and textual outputs. The final unified output representation $X_o$ is obtained by:
\begin{equation}
X_o = [X_m^{O}W_m^{O}, X_{t}^OW_t^{O}]
\end{equation}
ensuring separate processing streams for each modality within the model. Similarly, given input $X$ for each feed-forward layer, the output is:
\begin{equation}
\text{FFN}(X) = [\text{FFN}_{m}(X_m), \text{FFN}_{t}(X_t)]
\end{equation}
where $\text{FFN}_{m}$ and $\text{FFN}_{t}$ are feed-forward layers for modality-specific and text inputs, respectively.

When composing MLLMs that are trained through parameter decoupling, we merge only the text-related parameters, maintaining distinct modality-specific parameters as depicted in Figure \ref{fig:arch}(d). The composite model functions as a natural extension of the methodology previously described, guaranteeing that inputs from each modality are processed independently with their respective parameters. By doing so, it effectively mitigates the risk of interference from other modalities, ensuring that the composite model maintains high fidelity in processing multimodal data.
Please note that after the MLLMs are trained, the composition phase remains emphatically \textit{training-free}.

\subsection{Adaptive Parameter Adjustment}
\label{sec:adjust}

Due to variations in data quality and training strategies among different MLLMs, their performance can significantly differ. Thus, when composing these models, employing a simplistic averaging strategy often falls short of achieving optimal results. To enhance model effectiveness, it is crucial to implement adaptive adjustments to the model parameters during the composition process, allowing for better compatibility and flexibility among the different models. Specifically, for $N$ distinct MLLMs $\{M_1, M_2, \ldots, M_N\}$ and any parameter $\theta_i$ common across these models, where $i=1...N$, the merged parameter is defined as:
\begin{equation}
\theta_{\text{merge}} = \sum_{i=1}^{N}\lambda_i \theta_i
\end{equation}
where $\lambda_i$ represents the adjustment coefficient. For simplicity, we adopt a uniform adjustment coefficient $\lambda_i$ for all $\theta_i$ in $M_i$. For models trained using parameter decoupling, we can additionally adjust their modality-specific parameters if needed. The values of these coefficients can be determined with a validation set from target tasks requiring various modal inputs. If such a validation set is not available, a practical alternative is to select the coefficients based on general performance of the model on tasks of each modality.
We refer to the updated model composition framework with parameter decoupling and adjustment as DAMC.

\subsection{Multimodal Commonality Understanding Benchmark}
\label{sec:mcub}

\begin{figure}[t]
    \centering
    \includegraphics[width=1.0\linewidth]{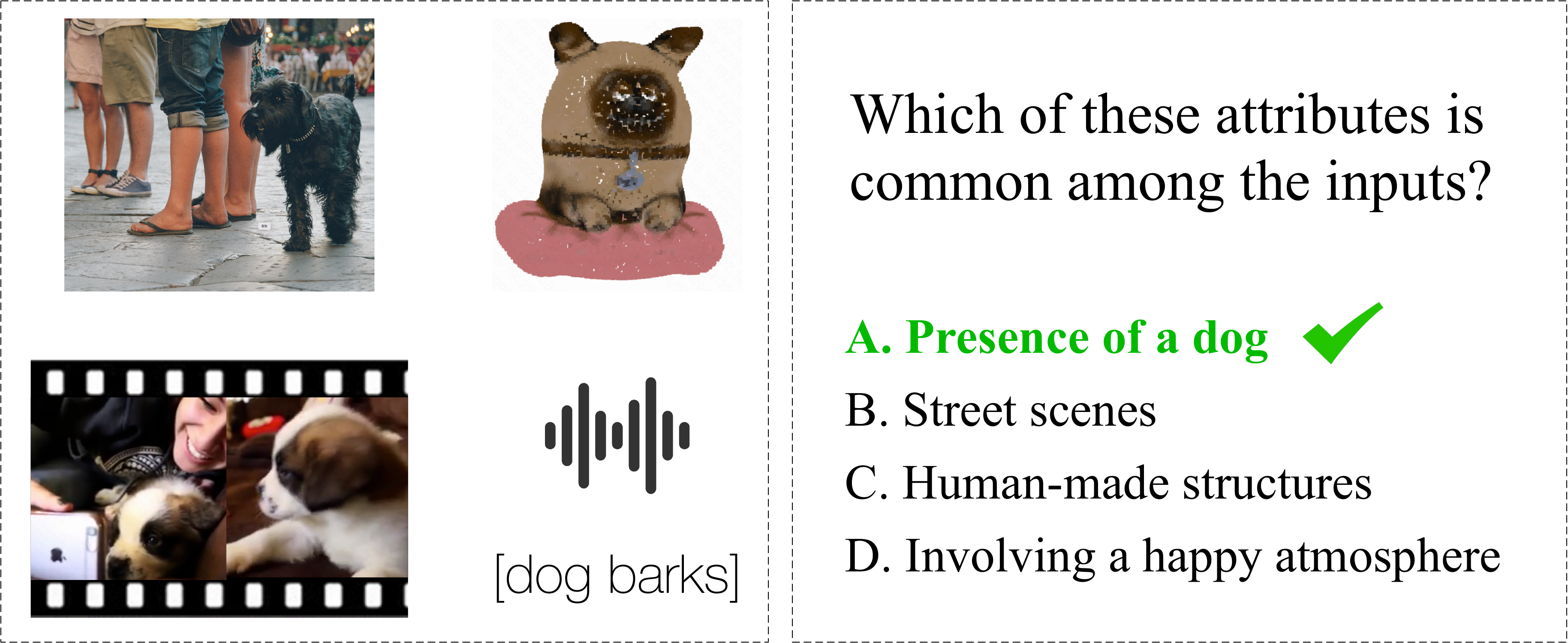}
    \caption{An example of MCUB-4, where the objective is to identify common attributes from inputs across four different modalities.}
    \label{fig:mcub}
\end{figure}

As previously discussed, DAMC inherently possesses the advantage of scaling up the number of modalities. However, there are few benchmarks available to evaluate the performance across various modalities. To demonstrate the effectiveness of our approach on tasks involving numerous modalities, inspired by \citet{panagopoulou2023xinstruct}, we introduce a new benchmark called the Multimodal Commonality Understanding Benchmark (MCUB). We provide an example of MCUB in Figure \ref{fig:mcub}. The task of MCUB is to measure the ability of the model to identify commonalities among input entities from diverse modalities and select the most appropriate answer from four given candidates.

We leverage GPT-4~\cite{achiam2023gpt4} to create MCUB from existing captioning datasets across various modalities. To be specific, we begin by randomly selecting groups of inputs from each modality, and retain the groups with the highest semantic similarity. This similarity is quantified by averaging the similarities of their respective captions. Subsequently, GPT-4 is prompted with in-context examples to generate questions, options, and correct answers for each group. In this study, we develop two variants of MCUB: MCUB-3 and MCUB-4. MCUB-4 comprises data entries that include inputs from four modalities—image, video, audio, and point cloud. In contrast, MCUB-3 consists of four subsets, each representing combinations of inputs from any three of these modalities. 
For more details, please refer to Appendix \ref{appendix:MCUB}.

\begin{table*}
\centering
\begin{tabular}{l|l|c|cc|c}
\toprule
\textbf{Task} & \textbf{Method} & \textbf{V} & \textbf{V + I} & \textbf{V + A} & \textbf{V + I + A} \\
\midrule
\multirow{7}{*}{MUSIC-AVQA} & ChatBridge-13B & - & - & 43.00 & - \\
& OneLLM-7B & - & - & 47.60 & - \\
& ImageBind-LLM & 37.24 & 38.76 & 39.72 & 38.16 \\
& X-InstructBLIP & 45.83 & 41.23 & 48.34 & 47.39 \\
\rowcolor{gray!10}
\multicolumn{1}{l|}{\cellcolor{white}} & Proj-only & 44.93 & 46.64  & 46.17  & 50.21  \\
\rowcolor{gray!10}
\multicolumn{1}{l|}{\cellcolor{white}} & \baseline & 49.00 & 52.52  & 50.66  & 53.63 \\
\rowcolor{gray!10}
\multicolumn{1}{l|}{\cellcolor{white}} & \ours  & \textbf{49.09} & \textbf{53.08}  & \textbf{50.91} & \textbf{57.32} \\
\midrule
\multirow{5}{*}{AVQA} & ImageBind-LLM & 51.77 & 51.65 & 55.00 & 54.26 \\
& X-InstructBLIP & 41.91 & 40.42 & 44.29 & 44.23 \\
\rowcolor{gray!10}
\multicolumn{1}{l|}{\cellcolor{white}AVQA} & Proj-only & 67.99 & 66.65 & 67.65 & 66.85 \\
\rowcolor{gray!10}
\multicolumn{1}{l|}{\cellcolor{white}} & \baseline & \textbf{79.37} & 79.74 & 79.82 & 80.70 \\
\rowcolor{gray!10}
\multicolumn{1}{l|}{\cellcolor{white}} & \ours & 79.15 & \textbf{80.30} & \textbf{80.40} & \textbf{81.31} \\
\bottomrule
\end{tabular}
\caption{Experimental results on zero-shot audio-visual question answering tasks with different combinations of video (V), image (I) and audio (A) inputs. Methods developed in this study are distinguished with a \colorbox{gray!10}{grey} background for clarity.}
\label{tab:main_avqa}
\end{table*}

\section{Experiments}
\label{sec:exp}

\subsection{Implementation}

In our experimental setup, we explore four modalities: \textit{image}, \textit{audio}, \textit{video} and \textit{point cloud}. To ensure a comprehensive and comparative analysis, we reimplement MLLMs for each modality following previous works~\cite{liu2023improved,lin2023videollava,panagopoulou2023xinstruct,xu2023pointllm}. For each of these modalities, we train three versions of MLLMs with same data and hyperparameters but vary the trainablity of LLM parameters: a model with a frozen LLM, a fully trainable LLM, and a trainable LLM with parameter decoupling, as illustrated in Figure \ref{fig:arch}. We employ the LoRA~\cite{hu2021lora} technique for efficient LLM training. All models are based on Vicuna-7B-v1.5~\cite{zheng2023judging}. 
More details for training the MLLMs are in Appendix \ref{appendix:pretrain_llms}. We apply parameter adjustment by conducting a search for the optimal $\lambda_i$ values within the set $\left[1/N, 2/N, \ldots, N/N\right]$ for composition of 
$N$ modalities based on the performance of the validation set for each corresponding task. The results for the best-performing parameters are documented in Appendix \ref{appendix:param_adjust}.

\subsection{Evaluation}

\paragraph{Datasets and Benchmarks.} Our aim is to evaluate the performace of the composite MLLM to process inputs from multiple modalities. In addition to the MCUB benchmark described in Section \ref{sec:mcub}, we also include datasets from two key areas for evaluation: (1) \textit{audio-visual question answering} including MUSIC-AVQA~\cite{li2022musicavqa} and AVQA~\cite{yang2022avqa} where image, video and audio inputs are available and (2) \textit{3D object classification} on ModelNet40~\cite{wu2015modelnet} and Objaverse~\cite{deitke2023objaverse} where image and point cloud inputs are considered. 

\paragraph{Baselines.} Our evaluation strategy involves comparison with baselines primarily focusing on models that do not leverage training data from multiple modalities. We consider two representative baselines for this purpose: \textbf{ImageBind-LLM} that aligns a multimodal encoder, ImageBind, to the LLM with image-text paired data and \textbf{X-InstructBLIP} that aligns individual modal encoders to a frozen LLM using instruction data specific to each modality.
Regarding model composition, we further compare our \ours{} approach against two alternative intermediate baselines as discussed before: \textbf{Proj-only} where the LLM is frozen and only the modal encoders and connectors are composed and \textbf{\baseline} that directly composes MLLMs without adopting our proposed parameter decoupling and adjustment techniques.

\begin{table}[ht]
\centering\resizebox{1.0\linewidth}{!}{
    \begin{tabular}{l|l|cc}
    \toprule
    \textbf{Task} & \textbf{Method} &  \textbf{Type-I} & \textbf{Type-C} \\
    \midrule
     & 3D-LLM & 49.00 & 41.50 \\
     & ImageBind-LLM & 31.00 & 26.50 \\
    \multicolumn{1}{l|}{\cellcolor{white}Objaverse} & X-InstructBLIP & 50.00 & 31.50 \\
     \rowcolor{gray!10}
     \multicolumn{1}{l|}{\cellcolor{white}} & Proj-only & 48.00 & 42.50 \\
     \rowcolor{gray!10}
     \multicolumn{1}{l|}{\cellcolor{white}} & \baseline{} & 55.00 & 59.50 \\
     \rowcolor{gray!10}
     \multicolumn{1}{l|}{\cellcolor{white}} & \ours{} & \textbf{60.50} & \textbf{62.00} \\
    \midrule
     & ImageBind-LLM & 42.71 & 42.46 \\
     & X-InstructBLIP & 61.43 & 61.14 \\
     \rowcolor{gray!10}
    \multicolumn{1}{l|}{\cellcolor{white}ModelNet40} & Proj-only & 62.88 & 61.99 \\
     \rowcolor{gray!10}
     \multicolumn{1}{l|}{\cellcolor{white}} & \baseline{} & 66.00 & 64.59 \\
     \rowcolor{gray!10}
     \multicolumn{1}{l|}{\cellcolor{white}} & \ours{} & \textbf{70.02} & \textbf{65.24} \\
    \bottomrule
    \end{tabular}}
    \caption{Experimental results on zero-shot 3D object classification tasks using combined point and image inputs (P + I). Following \citet{xu2023pointllm}, two different types of prompts are considered: Type-I (``What is this?'') and Type-C (``This is an object of'').}
    \label{tab:main_point}
\end{table}

\subsection{Main Results}

The main experimental results on audio-visual question answering, 3D object classification and MCUB are presented in Table \ref{tab:main_avqa}, Table \ref{tab:main_point} and Table \ref{tab:main_mcub}, respectively. 
In the tables, the notation ``X+Y'' represents the combination of different types of modal inputs used in evaluating the models. For example, ``V + A'' refers to the combination of video (V) and audio (A) inputs being used together as part of the input to the model. For model composition, we compose MLLMs based on the specific modal inputs required. Generally, our proposed \ours{} achieves the highest performance across all input combinations and tasks, demonstrating its effectiveness of adeptly managing inputs from multiple modalities even without any training on these modal combinations. In addition, we make the following observations:

\paragraph{Model composition brings performance improvements with additional modal inputs.} The results reveal a generally consistent trend where the application of model composition methods leads to improved performance as more modal inputs are integrated. For example, in Table \ref{tab:main_avqa} in the MUSIC-AVQA task, the transition from solely video (V) inputs to combinations that include image (I) and audio (A) inputs leads to noteworthy performance boosts. Specifically, when combining these three modalities, the performance for our \ours{} reaches a peak at 57.32, marking a notable increase compared to V+I/A. In contrast, models like ImageBind-LLM and X-InstructBLIP do not show similar improvements when transitioning from V+I/A to the V+I+A combination. This demonstrates that models developed through model composition can handle inputs from different modalities more effectively, showcasing their ability to improve multimodal understanding.

\begin{table}
\centering
\begin{tabular}{l|cc}
\toprule
Method & \benchmark{}-3 & \benchmark{}-4 \\
\midrule
ImageBind-LLM & 32.95 & 32.93 \\
X-InstructBLIP & 29.30 & 27.94 \\
\rowcolor{gray!10}
Proj-only & 44.15 & 43.00 \\
\rowcolor{gray!10}
\baseline & 54.70 & 54.03 \\
\rowcolor{gray!10}
\ours & \textbf{59.80} & \textbf{60.08} \\
\bottomrule
\end{tabular}
\caption{Results on \benchmark{}. \benchmark{}-3 refers to subsets of the data with inputs from three modalities, while \benchmark{}-4 includes inputs from four modalities.}
\label{tab:main_mcub}
\end{table}

\paragraph{\ours{} outperforms strong baselines.} Compared to previous methods, \ours{} achieves superior performance, even surpassing methods that utilize paired multimodal data, such as ChatBridge and 3D-LLM~\cite{hong20233dllm}.  When comparing the performance across various model composition methods, \ours{} significantly outperforms Proj-only and \baseline, particularly evident in scenarios involving the integration of more than two modalities. On the MCUB benchmark, for instance, \ours{} achieves a significant performance enhancement, with an improvement over the second best \baseline{} of +5.10 points in scenarios combining three modalities and +6.05 points in the subset involving four modalities. This distinction proves the capability of \ours{} to mitigate interference effectively when merging multiple MLLMs, demonstrating its robustness and efficiency for model composition.

\subsection{Ablation Study}

Our ablation study, summarized in Table \ref{tab:ablation}, evaluates the effects of parameter decoupling and adjustment on \ours{}. The findings reveal that employing neither strategy yields an average performance of 62.79 across benchmarks. Introducing parameter decoupling alone enhances the average to 65.05, while adjustment alone improves it to 63.55. Remarkably, combining both strategies boosts the average performance to 66.24, with significant improvements noted in all benchmarks. This highlights the synergistic impact of parameter decoupling and adjustment in optimizing multimodal model composition and achieving superior performance across varied tasks involving different types of modality inputs.

\subsection{Qualitative Results}

\begin{figure}[t]
\centering
\includegraphics[width=\linewidth]{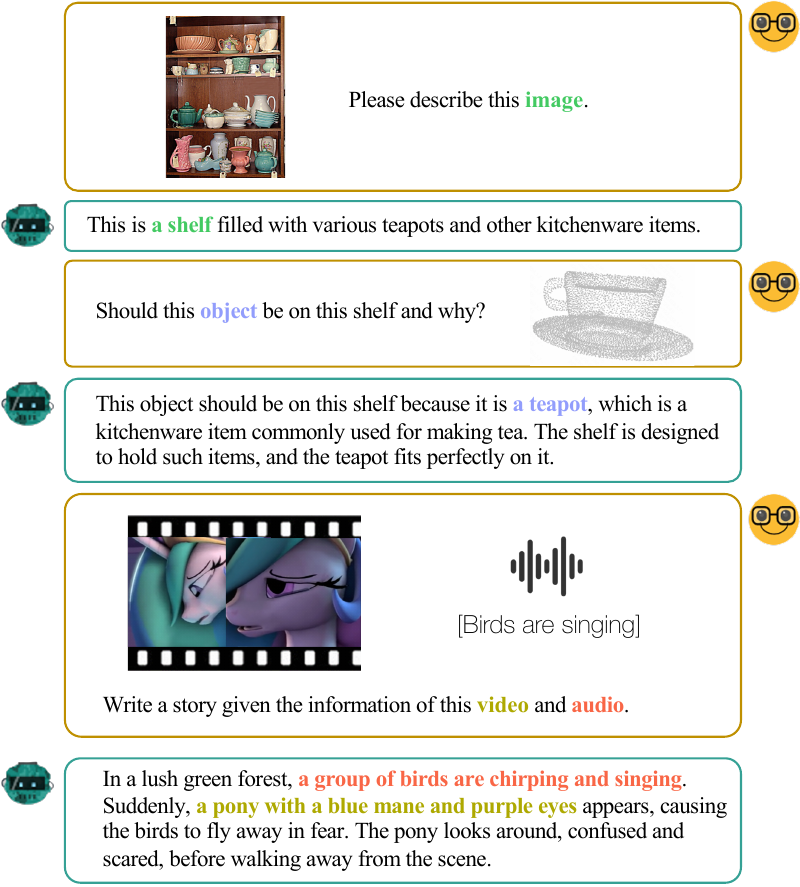}
\caption{Qualitative examples on multimodal understanding of a composite model integrating four MLLMs. }
\label{fig:qualitative}
\end{figure}

Figure \ref{fig:qualitative} demonstrate the capability of the composition model to understand and reason over multi-modality inputs. Additional qualitative results are in Appendix \ref{appendix:additional_qr}.

\section{Discussion and Analysis}

\begin{table*}[ht]
\centering
\begin{minipage}{0.60\textwidth}
\centering
\resizebox{\textwidth}{!}{%
\begin{tabular}{cc|cccc}
\toprule
Decoup. & Adjust. & MUSIC-AVQA & AVQA & \benchmark{}-4 & Avg. \\ 
\midrule
\xmark & \xmark & 53.63 & 80.70 & 54.03 & 62.79 \\
\cmark & \xmark & 53.94 & 81.12 & \textbf{60.08} & 65.05\\
\xmark & \cmark & 55.36 & 81.27 & 54.03 & 63.55\\
\midrule 
\cmark & \cmark & \textbf{57.32} & \textbf{81.31} & \textbf{60.08} & \textbf{66.24} \\
\bottomrule
\end{tabular}}
\caption{Ablation study on various components of \ours{}. ``Decoup.'' and ``Adjust.'' denote parameter decoupling and adjustment.}
\label{tab:ablation}
\end{minipage}
\hfill 
\begin{minipage}{0.375\textwidth}
\centering
\resizebox{\textwidth}{!}{%
\begin{tabular}{lccc}
\toprule
Modal & \baseline{} & \ours{} & $\Delta$ \\ \midrule
I    &  1780.32    &  1772.85  & -0.42\% \\
\midrule
A + I   &  1718.42    &  1729.72 & +0.66\%   \\
V + I   &  1780.16    &  1825.50 & +2.55\%   \\
\midrule
V + I + A  &  1709.14    & 1852.48 & +8.39\%    \\ \bottomrule
\end{tabular}}
\caption{Composite model performance on the MME benchmark.}
\label{tab:discussion_mme}
\end{minipage}
\end{table*}

\begin{table}[ht]
\centering
\begin{tabular}{cc|c}
\toprule
\multicolumn{2}{c|}{Parameter Decoupling}  & MUSIC-AVQA \\ 
V & I  & V+I+A \\ \midrule
No & No   & 55.36 \\ \midrule
Infer. & Train.       & 56.28 \\
Train. & Infer.       & 56.42 \\
Infer. & Infer.     & 56.16  \\ \midrule
Train. & Train.        & \textbf{57.32} \\ \bottomrule
\end{tabular}
\caption{Results of different parameter decoupling strategies. ``Infer.'' and ``Train.'' denote parameter decoupling at inference or training time.}
\label{tab:convert}
\end{table}

\paragraph{Composition of MLLMs with different training strategies.} In Section \ref{sec:pd}, we advocate training MLLMs with parameter decoupling. Intriguingly, this decoupling can also be implemented at inference time, by replicating the parameters to mimicking identical parameters on modality-specific and textual inputs. In this way, we can compose models with different training strategies. We conduct experiments on the MUSIC-AVQA dataset with V + I + A inputs. We vary the training strategy of video and image MLLMs, and compose them with an audio MLLMs trained with parameter decoupling. The results in Table \ref{tab:convert} indicate that decoupling at inference time is also effective, which is beneficial when aiming to integrate existing models that were not trained with parameter decoupling.

\paragraph{Composite model performance on the single-modality task.} To explore the effectiveness of the composite model on single-modality tasks, we conducted experiments on the vision-language understanding evaluation benchmark MME~\cite{fu2023mme}. The results are shown in Table \ref{tab:discussion_mme}. Each composite model processes only images and textual questions. The $\Delta$ values denote the relative improvements of \ours{} over \baseline{}. As more models are integrated, the performance of \baseline{} tends to decline, while \ours{} consistently shows improvement.
This highlights the capability of our proposed methods to mitigate parameter interference effectively when composing models. Moreover, it is noteworthy that model composition enhances performance even in tasks limited to a single modality.  We attribute this to the possibility that our approach could integrate knowledge from different models during the combination process, which we will leave for future research. 

\paragraph{Impact of trainable parameters.} We investigate whether the benefits of parameter decoupling arise from the augmentation of trainable parameters, given that two distinct sets of parameters are allocated for varying inputs. To this end, we train a new vision LLM without decoupling but increase the rank of the LoRA modules from 128 to 256, which doubles the trainable parameters. The findings, as shown in Table \ref{tab:lora2}, reveal that merely expanding the pool of trainable parameters does not enhance performance. This suggests that the advantage of parameter decoupling extends beyond simple parameter quantity increase, highlighting its effectiveness in mitigating parameter interference.

\begin{table}[ht]
\centering
\begin{tabular}{lcc}
\toprule
Method & MUSIC-AVQA & AVQA \\ \midrule
\ours{}\,\,\,\,\,\, (r=128) & 57.32 & 81.31 \\
\midrule
\baseline{} (r=128)    &  53.63    &  80.70   \\
\baseline{} (r=256)   &  52.79    & 80.81    \\
\bottomrule
\end{tabular}
\caption{Composite model performance on zero-shot audio-visual question answering tasks.}
\label{tab:lora2}
\end{table}

\section{Conclusion}

In this paper, we introduce and explore the concept of model composition for MLLMs, showcasing its ability to seamlessly integrate diverse modal capabilities without additional training. Initially, we introduce a foundational model composition framework, referred to as \baseline{}. Advancing further, we design \ours{} that employs parameter decoupling and adjustment to reduce cross-modal interference and optimize performance of the composite model. In addition, we construct a benchmark \benchmark{} for multimodal commonality understanding evaluation to facilitate the relative research. Extensive experiments demonstrate the effectiveness of our approaches. We hope that our work will inspire further exploration into model composition for multimodal models.

\section*{Limitations}

Our exploration is restricted to four commonly used modalities, omitting a comprehensive examination across the entire spectrum of potential modalities. This limitation may result in missed opportunities to identify further benefits or challenges associated with model composition. Additionally, our proposed approach has been tested on models of specific sizes, leaving the applicability of the mehods on larger-scale MLLMs an open question for future research.

\section*{Acknowledgments}

This work is supported by the National Key R\&D Program of China (2022ZD0160502) and the National Natural Science Foundation of China (No. 61925601, 62276152).

\bibliography{custom}

\begin{thebibliography}{44}
\expandafter\ifx\csname natexlab\endcsname\relax\def\natexlab#1{#1}\fi

\bibitem[{Achiam et~al.(2023)Achiam, Adler, Agarwal, Ahmad, Akkaya, Aleman, Almeida, Altenschmidt, Altman, Anadkat et~al.}]{achiam2023gpt4}
Josh Achiam, Steven Adler, Sandhini Agarwal, Lama Ahmad, Ilge Akkaya, Florencia~Leoni Aleman, Diogo Almeida, Janko Altenschmidt, Sam Altman, Shyamal Anadkat, et~al. 2023.
\newblock Gpt-4 technical report.
\newblock \emph{arXiv preprint arXiv:2303.08774}.

\bibitem[{Chen et~al.(2023)Chen, Han, Zhao, Zhang, Shi, Xu, and Xu}]{chen2023xllm}
Feilong Chen, Minglun Han, Haozhi Zhao, Qingyang Zhang, Jing Shi, Shuang Xu, and Bo~Xu. 2023.
\newblock \href {http://arxiv.org/abs/2305.04160} {X-llm: Bootstrapping advanced large language models by treating multi-modalities as foreign languages}.

\bibitem[{Chen et~al.(2022)Chen, Wu, Wang, Liu, Tompkins, Chen, and Wei}]{chen2022beats}
Sanyuan Chen, Yu~Wu, Chengyi Wang, Shujie Liu, Daniel Tompkins, Zhuo Chen, and Furu Wei. 2022.
\newblock Beats: Audio pre-training with acoustic tokenizers.
\newblock \emph{arXiv preprint arXiv:2212.09058}.

\bibitem[{Dai et~al.(2023)Dai, Li, Li, Tiong, Zhao, Wang, Li, Fung, and Hoi}]{dai2023instructblip}
Wenliang Dai, Junnan Li, Dongxu Li, Anthony Tiong, Junqi Zhao, Weisheng Wang, Boyang Li, Pascale Fung, and Steven Hoi. 2023.
\newblock \href {https://openreview.net/forum?id=vvoWPYqZJA} {Instruct{BLIP}: Towards general-purpose vision-language models with instruction tuning}.
\newblock In \emph{Thirty-seventh Conference on Neural Information Processing Systems}.

\bibitem[{Deitke et~al.(2023)Deitke, Schwenk, Salvador, Weihs, Michel, VanderBilt, Schmidt, Ehsani, Kembhavi, and Farhadi}]{deitke2023objaverse}
Matt Deitke, Dustin Schwenk, Jordi Salvador, Luca Weihs, Oscar Michel, Eli VanderBilt, Ludwig Schmidt, Kiana Ehsani, Aniruddha Kembhavi, and Ali Farhadi. 2023.
\newblock Objaverse: A universe of annotated 3d objects.
\newblock In \emph{Proceedings of the IEEE/CVF Conference on Computer Vision and Pattern Recognition}, pages 13142--13153.

\bibitem[{Deshmukh et~al.(2023)Deshmukh, Elizalde, Singh, and Wang}]{deshmukh2023pengi}
Soham Deshmukh, Benjamin Elizalde, Rita Singh, and Huaming Wang. 2023.
\newblock Pengi: An audio language model for audio tasks.
\newblock \emph{arXiv preprint arXiv:2305.11834}.

\bibitem[{Fu et~al.(2023)Fu, Chen, Shen, Qin, Zhang, Lin, Yang, Zheng, Li, Sun et~al.}]{fu2023mme}
Chaoyou Fu, Peixian Chen, Yunhang Shen, Yulei Qin, Mengdan Zhang, Xu~Lin, Jinrui Yang, Xiawu Zheng, Ke~Li, Xing Sun, et~al. 2023.
\newblock Mme: A comprehensive evaluation benchmark for multimodal large language models.
\newblock \emph{arXiv preprint arXiv:2306.13394}.

\bibitem[{Girdhar et~al.(2023)Girdhar, El-Nouby, Liu, Singh, Alwala, Joulin, and Misra}]{girdhar2023imagebind}
Rohit Girdhar, Alaaeldin El-Nouby, Zhuang Liu, Mannat Singh, Kalyan~Vasudev Alwala, Armand Joulin, and Ishan Misra. 2023.
\newblock Imagebind: One embedding space to bind them all.
\newblock In \emph{Proceedings of the IEEE/CVF Conference on Computer Vision and Pattern Recognition}, pages 15180--15190.

\bibitem[{Gong et~al.(2023)Gong, Luo, Liu, Karlinsky, and Glass}]{gong2023ltu}
Yuan Gong, Hongyin Luo, Alexander~H Liu, Leonid Karlinsky, and James Glass. 2023.
\newblock Listen, think, and understand.
\newblock \emph{arXiv preprint arXiv:2305.10790}.

\bibitem[{Han et~al.(2023)Han, Zhang, Shao, Gao, Xu, Xiao, Zhang, Liu, Wen, Guo et~al.}]{han2023imagebindllm}
Jiaming Han, Renrui Zhang, Wenqi Shao, Peng Gao, Peng Xu, Han Xiao, Kaipeng Zhang, Chris Liu, Song Wen, Ziyu Guo, et~al. 2023.
\newblock Imagebind-llm: Multi-modality instruction tuning.
\newblock \emph{arXiv preprint arXiv:2309.03905}.

\bibitem[{Hong et~al.(2023)Hong, Zhen, Chen, Zheng, Du, Chen, and Gan}]{hong20233dllm}
Yining Hong, Haoyu Zhen, Peihao Chen, Shuhong Zheng, Yilun Du, Zhenfang Chen, and Chuang Gan. 2023.
\newblock 3d-llm: Injecting the 3d world into large language models.
\newblock \emph{arXiv preprint arXiv:2307.12981}.

\bibitem[{Hu et~al.(2021)Hu, Shen, Wallis, Allen-Zhu, Li, Wang, Wang, and Chen}]{hu2021lora}
Edward~J Hu, Yelong Shen, Phillip Wallis, Zeyuan Allen-Zhu, Yuanzhi Li, Shean Wang, Lu~Wang, and Weizhu Chen. 2021.
\newblock Lora: Low-rank adaptation of large language models.
\newblock \emph{arXiv preprint arXiv:2106.09685}.

\bibitem[{Huang et~al.(2023)Huang, Liu, Lin, Pang, Du, and Lin}]{huang2023lorahub}
Chengsong Huang, Qian Liu, Bill~Yuchen Lin, Tianyu Pang, Chao Du, and Min Lin. 2023.
\newblock Lorahub: Efficient cross-task generalization via dynamic lora composition.
\newblock \emph{arXiv preprint arXiv:2307.13269}.

\bibitem[{Ilharco et~al.(2022)Ilharco, Ribeiro, Wortsman, Schmidt, Hajishirzi, and Farhadi}]{ilharco2022taskarith}
Gabriel Ilharco, Marco~Tulio Ribeiro, Mitchell Wortsman, Ludwig Schmidt, Hannaneh Hajishirzi, and Ali Farhadi. 2022.
\newblock Editing models with task arithmetic.
\newblock In \emph{The Eleventh International Conference on Learning Representations}.

\bibitem[{Jin et~al.(2023)Jin, Ren, Preotiuc-Pietro, and Cheng}]{jin2023dataless}
Xisen Jin, Xiang Ren, Daniel Preotiuc-Pietro, and Pengxiang Cheng. 2023.
\newblock \href {http://arxiv.org/abs/2212.09849} {Dataless knowledge fusion by merging weights of language models}.

\bibitem[{Kim et~al.(2019)Kim, Kim, Lee, and Kim}]{audiocaps}
Chris~Dongjoo Kim, Byeongchang Kim, Hyunmin Lee, and Gunhee Kim. 2019.
\newblock Audiocaps: Generating captions for audios in the wild.
\newblock In \emph{NAACL-HLT}.

\bibitem[{Li et~al.(2022)Li, Wei, Tian, Xu, Wen, and Hu}]{li2022musicavqa}
Guangyao Li, Yake Wei, Yapeng Tian, Chenliang Xu, Ji-Rong Wen, and Di~Hu. 2022.
\newblock Learning to answer questions in dynamic audio-visual scenarios.
\newblock In \emph{Proceedings of the IEEE/CVF Conference on Computer Vision and Pattern Recognition}, pages 19108--19118.

\bibitem[{Lin et~al.(2023)Lin, Zhu, Ye, Ning, Jin, and Yuan}]{lin2023videollava}
Bin Lin, Bin Zhu, Yang Ye, Munan Ning, Peng Jin, and Li~Yuan. 2023.
\newblock Video-llava: Learning united visual representation by alignment before projection.
\newblock \emph{arXiv preprint arXiv:2311.10122}.

\bibitem[{Lin et~al.(2014)Lin, Maire, Belongie, Hays, Perona, Ramanan, Doll{\'a}r, and Zitnick}]{lin2014microsoft}
Tsung-Yi Lin, Michael Maire, Serge Belongie, James Hays, Pietro Perona, Deva Ramanan, Piotr Doll{\'a}r, and C~Lawrence Zitnick. 2014.
\newblock Microsoft coco: Common objects in context.
\newblock In \emph{Computer Vision--ECCV 2014: 13th European Conference, Zurich, Switzerland, September 6-12, 2014, Proceedings, Part V 13}, pages 740--755. Springer.

\bibitem[{Liu et~al.(2023{\natexlab{a}})Liu, Li, Li, and Lee}]{liu2023improved}
Haotian Liu, Chunyuan Li, Yuheng Li, and Yong~Jae Lee. 2023{\natexlab{a}}.
\newblock Improved baselines with visual instruction tuning.
\newblock \emph{arXiv preprint arXiv:2310.03744}.

\bibitem[{Liu et~al.(2023{\natexlab{b}})Liu, Li, Wu, and Lee}]{liu2023visual}
Haotian Liu, Chunyuan Li, Qingyang Wu, and Yong~Jae Lee. 2023{\natexlab{b}}.
\newblock Visual instruction tuning.
\newblock \emph{arXiv preprint arXiv:2304.08485}.

\bibitem[{Luo et~al.(2023)Luo, Rockwell, Lee, and Johnson}]{luo2023scalable}
Tiange Luo, Chris Rockwell, Honglak Lee, and Justin Johnson. 2023.
\newblock \href {http://arxiv.org/abs/2306.07279} {Scalable 3d captioning with pretrained models}.

\bibitem[{Lyu et~al.(2023)Lyu, Wu, Wang, Huang, Liu, Du, Shi, and Tu}]{lyu2023macaw}
Chenyang Lyu, Minghao Wu, Longyue Wang, Xinting Huang, Bingshuai Liu, Zefeng Du, Shuming Shi, and Zhaopeng Tu. 2023.
\newblock Macaw-llm: Multi-modal language modeling with image, audio, video, and text integration.
\newblock \emph{arXiv preprint arXiv:2306.09093}.

\bibitem[{Maaz et~al.(2023)Maaz, Rasheed, Khan, and Khan}]{maaz2023videochatgpt}
Muhammad Maaz, Hanoona Rasheed, Salman Khan, and Fahad~Shahbaz Khan. 2023.
\newblock Video-chatgpt: Towards detailed video understanding via large vision and language models.
\newblock \emph{arXiv preprint arXiv:2306.05424}.

\bibitem[{Matena and Raffel(2022)}]{matena2022merging}
Michael Matena and Colin Raffel. 2022.
\newblock \href {http://arxiv.org/abs/2111.09832} {Merging models with fisher-weighted averaging}.

\bibitem[{Mei et~al.(2023)Mei, Meng, Liu, Kong, Ko, Zhao, Plumbley, Zou, and Wang}]{mei2023wavcaps}
Xinhao Mei, Chutong Meng, Haohe Liu, Qiuqiang Kong, Tom Ko, Chengqi Zhao, Mark~D Plumbley, Yuexian Zou, and Wenwu Wang. 2023.
\newblock Wavcaps: A chatgpt-assisted weakly-labelled audio captioning dataset for audio-language multimodal research.
\newblock \emph{arXiv preprint arXiv:2303.17395}.

\bibitem[{Panagopoulou et~al.(2023)Panagopoulou, Xue, Yu, Li, Li, Joty, Xu, Savarese, Xiong, and Niebles}]{panagopoulou2023xinstruct}
Artemis Panagopoulou, Le~Xue, Ning Yu, Junnan Li, Dongxu Li, Shafiq Joty, Ran Xu, Silvio Savarese, Caiming Xiong, and Juan~Carlos Niebles. 2023.
\newblock X-instructblip: A framework for aligning x-modal instruction-aware representations to llms and emergent cross-modal reasoning.
\newblock \emph{arXiv preprint arXiv:2311.18799}.

\bibitem[{Reimers and Gurevych(2019)}]{reimers2019sentencebert}
Nils Reimers and Iryna Gurevych. 2019.
\newblock \href {http://arxiv.org/abs/1908.10084} {Sentence-bert: Sentence embeddings using siamese bert-networks}.

\bibitem[{Su et~al.(2023)Su, Lan, Li, Xu, Wang, and Cai}]{su2023pandagpt}
Yixuan Su, Tian Lan, Huayang Li, Jialu Xu, Yan Wang, and Deng Cai. 2023.
\newblock Pandagpt: One model to instruction-follow them all.
\newblock \emph{arXiv preprint arXiv:2305.16355}.

\bibitem[{Sundar et~al.(2023)Sundar, Yang, Chan, Ghosh, Ravichandran, and Nidadavolu}]{sundar2023multimodal}
Anirudh~S Sundar, Chao-Han~Huck Yang, David~M Chan, Shalini Ghosh, Venkatesh Ravichandran, and Phani~Sankar Nidadavolu. 2023.
\newblock Multimodal attention merging for improved speech recognition and audio event classification.
\newblock \emph{arXiv preprint arXiv:2312.14378}.

\bibitem[{Sung et~al.(2023)Sung, Li, Lin, Gan, Bansal, and Wang}]{sung2023empirical}
Yi-Lin Sung, Linjie Li, Kevin Lin, Zhe Gan, Mohit Bansal, and Lijuan Wang. 2023.
\newblock An empirical study of multimodal model merging.
\newblock \emph{arXiv preprint arXiv:2304.14933}.

\bibitem[{Tang et~al.(2024)Tang, Yu, Sun, Chen, Tan, Li, Lu, MA, and Zhang}]{tang2024salmonn}
Changli Tang, Wenyi Yu, Guangzhi Sun, Xianzhao Chen, Tian Tan, Wei Li, Lu~Lu, Zejun MA, and Chao Zhang. 2024.
\newblock \href {https://openreview.net/forum?id=14rn7HpKVk} {{SALMONN}: Towards generic hearing abilities for large language models}.
\newblock In \emph{The Twelfth International Conference on Learning Representations}.

\bibitem[{Wortsman et~al.(2022)Wortsman, Ilharco, Kim, Li, Kornblith, Roelofs, Gontijo-Lopes, Hajishirzi, Farhadi, Namkoong, and Schmidt}]{wortsman2022robust}
Mitchell Wortsman, Gabriel Ilharco, Jong~Wook Kim, Mike Li, Simon Kornblith, Rebecca Roelofs, Raphael Gontijo-Lopes, Hannaneh Hajishirzi, Ali Farhadi, Hongseok Namkoong, and Ludwig Schmidt. 2022.
\newblock \href {http://arxiv.org/abs/2109.01903} {Robust fine-tuning of zero-shot models}.

\bibitem[{Wu et~al.(2023)Wu, Fei, Qu, Ji, and Chua}]{wu2023nextgpt}
Shengqiong Wu, Hao Fei, Leigang Qu, Wei Ji, and Tat-Seng Chua. 2023.
\newblock \href {http://arxiv.org/abs/2309.05519} {Next-gpt: Any-to-any multimodal llm}.

\bibitem[{Wu et~al.(2015)Wu, Song, Khosla, Yu, Zhang, Tang, and Xiao}]{wu2015modelnet}
Zhirong Wu, Shuran Song, Aditya Khosla, Fisher Yu, Linguang Zhang, Xiaoou Tang, and Jianxiong Xiao. 2015.
\newblock 3d shapenets: A deep representation for volumetric shapes.
\newblock In \emph{Proceedings of the IEEE conference on computer vision and pattern recognition}, pages 1912--1920.

\bibitem[{Xu et~al.(2016)Xu, Mei, Yao, and Rui}]{7780940}
Jun Xu, Tao Mei, Ting Yao, and Yong Rui. 2016.
\newblock \href {https://doi.org/10.1109/CVPR.2016.571} {Msr-vtt: A large video description dataset for bridging video and language}.
\newblock In \emph{2016 IEEE Conference on Computer Vision and Pattern Recognition (CVPR)}, pages 5288--5296.

\bibitem[{Xu et~al.(2023)Xu, Wang, Wang, Chen, Pang, and Lin}]{xu2023pointllm}
Runsen Xu, Xiaolong Wang, Tai Wang, Yilun Chen, Jiangmiao Pang, and Dahua Lin. 2023.
\newblock Pointllm: Empowering large language models to understand point clouds.
\newblock \emph{arXiv preprint arXiv:2308.16911}.

\bibitem[{Yang et~al.(2022)Yang, Wang, Duan, Chen, Hou, Jin, and Zhu}]{yang2022avqa}
Pinci Yang, Xin Wang, Xuguang Duan, Hong Chen, Runze Hou, Cong Jin, and Wenwu Zhu. 2022.
\newblock Avqa: A dataset for audio-visual question answering on videos.
\newblock In \emph{Proceedings of the 30th ACM International Conference on Multimedia}, pages 3480--3491.

\bibitem[{Yu et~al.(2024)Yu, Yu, Yu, Huang, and Li}]{yu2024language}
Le~Yu, Bowen Yu, Haiyang Yu, Fei Huang, and Yongbin Li. 2024.
\newblock \href {http://arxiv.org/abs/2311.03099} {Language models are super mario: Absorbing abilities from homologous models as a free lunch}.

\bibitem[{Zeng et~al.(2023)Zeng, Zhang, Zheng, Xia, Wei, Wei, Zhang, and Kong}]{zeng2023lynx}
Yan Zeng, Hanbo Zhang, Jiani Zheng, Jiangnan Xia, Guoqiang Wei, Yang Wei, Yuchen Zhang, and Tao Kong. 2023.
\newblock What matters in training a gpt4-style language model with multimodal inputs?
\newblock \emph{arXiv preprint arXiv:2307.02469}.

\bibitem[{Zhang et~al.(2023)Zhang, Li, and Bing}]{zhang2023videollama}
Hang Zhang, Xin Li, and Lidong Bing. 2023.
\newblock Video-llama: An instruction-tuned audio-visual language model for video understanding.
\newblock \emph{arXiv preprint arXiv:2306.02858}.

\bibitem[{Zhao et~al.(2023)Zhao, Guo, Yue, Chen, Shao, Zhu, Yuan, and Liu}]{zhao2023chatbridge}
Zijia Zhao, Longteng Guo, Tongtian Yue, Sihan Chen, Shuai Shao, Xinxin Zhu, Zehuan Yuan, and Jing Liu. 2023.
\newblock Chatbridge: Bridging modalities with large language model as a language catalyst.
\newblock \emph{arXiv preprint arXiv:2305.16103}.

\bibitem[{Zheng et~al.(2023)Zheng, Chiang, Sheng, Zhuang, Wu, Zhuang, Lin, Li, Li, Xing, Zhang, Gonzalez, and Stoica}]{zheng2023judging}
Lianmin Zheng, Wei-Lin Chiang, Ying Sheng, Siyuan Zhuang, Zhanghao Wu, Yonghao Zhuang, Zi~Lin, Zhuohan Li, Dacheng Li, Eric.~P Xing, Hao Zhang, Joseph~E. Gonzalez, and Ion Stoica. 2023.
\newblock \href {http://arxiv.org/abs/2306.05685} {Judging llm-as-a-judge with mt-bench and chatbot arena}.

\bibitem[{Zhu et~al.(2023)Zhu, Lin, Ning, Yan, Cui, Wang, Pang, Jiang, Zhang, Li et~al.}]{zhu2023languagebind}
Bin Zhu, Bin Lin, Munan Ning, Yang Yan, Jiaxi Cui, HongFa Wang, Yatian Pang, Wenhao Jiang, Junwu Zhang, Zongwei Li, et~al. 2023.
\newblock Languagebind: Extending video-language pretraining to n-modality by language-based semantic alignment.
\newblock \emph{arXiv preprint arXiv:2310.01852}.

\end{thebibliography}

\clearpage
\appendix

\section{Implementation Details of Pre-trained MLLMs}
\label{appendix:pretrain_llms}

Table \ref{tab:mlm_structure} details the components and training data of each MLLM across the four modalities, with specific explanations provided below: 

\begin{itemize}
    \item \textbf{Image}: We follow LLaVA-1.5~\cite{liu2023improved} to use CLIP-ViT-L-336px as the image encoder with an MLP projection as the connector. The models are trained in a two-stage manner with LCS 558K~\cite{liu2023visual} as stage-1 data and LLaVA-mixed 665K~\cite{liu2023improved} as stage-2 data.
    \item \textbf{Audio}: We use BEATs-Iter3+~\cite{chen2022beats} as the audio encoder and a Q-Former with 32 query tokens as the connector following X-InstructBLIP~\cite{panagopoulou2023xinstruct}. We use WaveCaps~\cite{mei2023wavcaps} for stage-1. For stage-2, we use a filtered version of OpenAQA~\cite{gong2023ltu} with 350K examples.
    \item \textbf{Video}: Following Video-LLaVA~\cite{lin2023videollava}, we use LanguageBind~\cite{zhu2023languagebind} as the video encoder with an MLP connector. We reuse the stage-1 weights from Video-LLaVA for resource saving. The stage-2 data comprises Video-ChatGPT~\cite{maaz2023videochatgpt} and a subset of LLaVA-mixed 665K including 100K image-text and 40K text-only instruction data. 
    \item \textbf{Point cloud}: We use the pre-trained point encoder and instruction-following data from PointLLM~\cite{xu2023pointllm}. We use an MLP projection as the connector.
\end{itemize}

We adopt the same hyperparameters mainly following previous works~\cite{liu2023improved, panagopoulou2023xinstruct,lin2023videollava,xu2023pointllm}, as listed in Table \ref{tab:hparam}. For the first training stage, only the parameters in the connectors are trainable. During the second training stage, for \ours{}, since we decouple the parameters of modality inputs and text inputs, we additionally adjust the learning rate for the text components to 2e-5 for all modalities. We apply the LoRA across all linear modules within the LLM, setting the LoRA rank to 128 and the alpha parameter to 256. For efficiency in training, we utilize DeepSpeed Zero Optimization stage 3.

\begin{table}[ht]
\centering
\begin{tabular}{lcccc}
\toprule
Task & Image & Audio & Video & PC  \\ \midrule
MUSIC-AVQA & 1 & 1/3 & 2/3 & - \\
AVQA & 1/3 & 2/3 & 2/3 & - \\
MCUB-4 & 1/4 & 1/4 & 1/4 & 1/4 \\
\bottomrule
\end{tabular}
\caption{Parameter adjustment weights for different MLLMs.}
\label{tab:adjust_weight}
\end{table}

\begin{table*}[ht]
\centering
\begin{tabular}{
  @{}
  >{\raggedright\arraybackslash}p{1.0cm}
  >{\raggedright\arraybackslash}p{3.0cm}
  >{\raggedright\arraybackslash}p{2cm}
  >{\raggedright\arraybackslash}p{3.5cm}
  >{\raggedright\arraybackslash}p{4.5cm}
  @{}
}
\toprule
Modal & Modal Encoder & Connector & Stage-1 Data & Stage-2 Data \\ \midrule
Image & CLIP-ViT-L-336px & MLP & LCS 558K & LLaVA-mixed 665K \\
\midrule
Audio & BEATs-Iter3+ & Q-Former & WaveCaps 400K & OpenAQA filtered 350K \\
\midrule
Video & LanguageBind & MLP & LCS 558K,\quad\quad\quad Valley 702K & Video-ChatGPT 100K, LLaVA-mixed sampled 140K \\
\midrule
Point Cloud & Point Encoder & MLP & PointLLM brief description 660K  & Point complex instruction 70K \\ \bottomrule
\end{tabular}
\caption{Components and training data of MLLMs for different modalities.}
\label{tab:mlm_structure}
\end{table*}

\begin{table*}[ht]
\centering
\begin{tabular}{llcccc}
\toprule
Stage & Hyperparameter & Image & Audio & Video & Point Cloud  \\ \midrule
\multirow{4}{*}{State-1} & Batch size & 256 & 256 & 256 & 128 \\
                          & LR & 1e-3 & 1e-3 & 1e-3 & 2e-3 \\
                          & LR Schedule & \multicolumn{4}{c}{cosine decay} \\
                          & Warmup Ratio & \multicolumn{4}{c}{0.03} \\
                          & Epoch & 1 & 1 & 1 & 3 \\
\midrule
\multirow{4}{*}{Stage-2} & Batch size & 128 & 128 & 128 & 32 \\
                          & LR & 2e-4 & 1e-4 & 2e-4 & 2e-5 \\
                          & LR Schedule & \multicolumn{4}{c}{cosine decay} \\
                          & Warmup Ratio & \multicolumn{4}{c}{0.03} \\
                          & Epoch & 1 & 1 & 1 & 3 \\
\bottomrule
\end{tabular}
\caption{Hyperparameters of different MLLMs.}
\label{tab:hparam}
\end{table*}

\section{Parameter Adjustment}
\label{appendix:param_adjust}

In preliminary experiments, we find that composing MLLMs rom two modalities typically achieved optimal results through a direct average, specifically a 1/2 + 1/2 combination. Due to time and resource constraints, we only conducted parameter adjustments on three types of tasks. This process involved conducting a search for the optimal $\lambda_i$ values within the set $\left[1/N, 2/N, \ldots, N/N\right]$ for the composition of 
$N$ modalities, based on the validation set performance for each corresponding task. The results are showcased in Table \ref{tab:adjust_weight}. We assume that the variance in coefficients comes from the differing demands of each task on the understanding capabilities across modalities. For MCUB-4, which requires a comprehensive grasp of content from all four modalities, an average coefficient emerged as the best result. Based on these findings, we also applied average coefficients for all MCUB-3 tasks.

\section{Details of Multimodal Commonality Understanding Benchmark}
\label{appendix:MCUB}

\begin{table}[ht]
\centering\small
\begin{tabular}{lc}
\toprule
Modal & Captioning Dataset  \\ \midrule
Image    &  COCO2017~\cite{lin2014microsoft} val set  \\
Video   &  MSRVTT~\cite{7780940} test set   \\
Audio   &   AudioCaps~\cite{audiocaps} test set   \\
Point Cloud  & Cap3D~\cite{luo2023scalable} (3000 subset)    \\ \bottomrule
\end{tabular}
\caption{Captioning datasets in each modality to generate MCUB benchmark.}
\label{tab:MCUB-Datasets}
\end{table}

Table \ref{tab:MCUB-Datasets} presents the captioning datasets we use to generate MCUB task. For point cloud modality, we reserve 3000 point clouds from Objaverse~\cite{deitke2023objaverse} dataset following \citet{xu2023pointllm}, and obtain their captions in Cap3D~\cite{luo2023scalable} dataset. Note that this part of the point clouds are \textbf{not} used in training and 3D object classification.

The semantic similarity of a group of entities is obtained by averaging the similarities of the captions of all two-entities combination, which are calculated by the \texttt{all-MiniLM-L6-v2} model~\cite{reimers2019sentencebert}. For example, if a group of entities with captions pair $(A, B, C)$ is provided, the group similarity of it will be the average of the similarities between $(A,B)$ , $(A,C)$ and $(B,C)$.

We report the detailed results on subtasks of MCUB-3 benchmark in Table \ref{tab:MCUB-3}. The final result of MCUB-3 is reported as the average of the four sub-tasks.

Prompt template for generate questions, options and correct answers: 

\begin{tcolorbox}[breakable,title=Prompt Template]
\small

\texttt{Given entity A with caption "A cat meowing and humans speaking on the background." with properties: agile, independent, and domesticated, entity B with caption "Loud barking and traffic" with properties aggressive, loud, and high energy, entity C with caption "Birds chirping in a quiet forest" and properties peaceful, wild, and vocal, and entity D with caption "A bustling city street with honking cars" and properties busy, noisy, and chaotic, you can generate a set of instruction answer pairs to find the commond point of the entities as follows:} \\
\texttt{Example: Question: Which of the followings are the common point of the four entities. A. Rhythmic ocean waves crashing on the shore. B. Gentle rustling of leaves in a serene garden. C. Audible environmental sounds. D. Soft crackling of a campfire under a starry sky.  Answer: C. Explanation: The maximum common point among the four entities is the presence of ambient environmental sounds, which can be perceived audibly. } \\
\texttt{Generate three such Question, Answer, Explanation triplets for entity A with caption "}\textit{<Caption A>}\texttt{" and properties "}\textit{<Properties A>}\texttt{", entity B with caption "}\textit{<Caption B>}\texttt{" and properties "}\textit{<Properties B>}\texttt{", entity C with caption "}\textit{<Caption C>}\texttt{" and properties "}\textit{<Properties C>}\texttt{", and entity D with caption "}\textit{<Caption D>}\texttt{" and properties "}\textit{<Properties D>}\texttt{"
Examples:}

\end{tcolorbox}

\begin{table}[ht]
\centering
\resizebox{0.95\linewidth}{!}{
\begin{tabular}{lcccc}
\toprule
\multirow{2}{*}{Method} & \multicolumn{4}{c}{MCUB-3} \\
             & V+I+A & V+A+P & V+I+P & I+A+P \\ \midrule
Imagebind-LLM & 35.20 & 31.40 & 31.80 & 33.40 \\
X-InstructBLIP & 41.40 & 25.20 & 29.40 & 21.20 \\
\rowcolor{gray!10}
Proj-only  & 47.40 & 43.80 & 42.60 & 42.80   \\ 
\rowcolor{gray!10}
NaiveMC  & 56.00 & 51.00 & 53.00 & 58.80   \\ 
\rowcolor{gray!10}
DAMC  & \textbf{56.60} & \textbf{58.80} & \textbf{58.20} & \textbf{65.60}   \\ \bottomrule
\end{tabular}}
\caption{Detailed results on sub-tasks of MCUB-3 involving different modalities combinations.}
\label{tab:MCUB-3}
\end{table}

\section{Prompt for Evaluation}

We list the evaluation prompts for each dataset and modality combination in Table \ref{tab:prompt_eval}. In the prompts, we use ``<image>'', ``<audio>'', ``<video>'' and ``<point>'' to represent image, audio, video and point cloud modality inputs.

\begin{table*}[ht]
\centering
\begin{tabular}{
  >{\raggedright\arraybackslash}p{2.5cm}
  >{\raggedright\arraybackslash}p{2.0cm}
  >{\raggedright\arraybackslash}p{9.5cm}
}
\toprule
Dataset & Modal & Prompt Template \\ \midrule
  & V & <video>\textbackslash n\{Question\} \textbackslash nAnswer the question using a single word.    \\
MUSIC-AVQA \& AVQA & V+I & Based on the video <video> and image <image>\textbackslash n\{Question\} \textbackslash nAnswer the question using a single word. \\
& V+A & Based on the video <video> and audio <audio>\textbackslash n\{Question\} \textbackslash nAnswer the question using a single word. \\ 
& V+I+A & Video: <video>\textbackslash n Image: <image>\textbackslash n Audio: <audio>\textbackslash n \{Question\} \textbackslash nAnswer the question using a single word. \\
\midrule
Objaverse \& ModelNet40 & P & <point>\textbackslash nWhat is this? (Type-I) / This is an object of (Type-C)     \\
& P+I & Based on rendered image <image> and point cloud <point>\textbackslash nWhat is this? (Type-I) / This is an object of (Type-C)\\
\midrule
  & V+I+A & Based on four input entities:\textbackslash nimage <image>\textbackslash naudio <audio>\textbackslash nvideo <video>\textbackslash n \{Question\} \{Options\} Answer with the option's letter from the given choices directly.     \\
MCUB-3 & V+A+P & Based on four input entities:\textbackslash naudio <audio>\textbackslash nvideo <video>\textbackslash npoint <point>\textbackslash n \{Question\} \{Options\} Answer with the option's letter from the given choices directly. \\
& V+I+P & Based on four input entities:\textbackslash nimage <image>\textbackslash nvideo <video>\textbackslash npoint <point>\textbackslash n \{Question\} \{Options\} Answer with the option's letter from the given choices directly. \\
& I+A+P & Based on three input entities:\textbackslash nimage <image>\textbackslash naudio <audio>\textbackslash npoint <point>\textbackslash n \{Question\} \{Options\} Answer with the option's letter from the given choices directly. \\ \midrule
MCUB-4& V+I+A+P &  Based on four input entities:\textbackslash nimage <image>\textbackslash naudio <audio>\textbackslash nvideo <video>\textbackslash npoint <point>\textbackslash n \{Question\} \{Options\} Answer with the option's letter from the given choices directly. \\ \bottomrule
\end{tabular}
\caption{Prompt Template for different evaluation benchmarks.}
\label{tab:prompt_eval}
\end{table*}

\section{Additional Point Cloud Results}

In Table \ref{tab:main_point}, we report exclusively on the zero-shot 3D object classification performance using P + I (Point cloud + Image) inputs, with comprehensive results detailed in Table \ref{tab:full_point}. We find that ImageBind-LLM, X-InstructBLIP, and Proj-only struggle with point cloud inputs alone, particularly in adhering to open-ended generation instructions, leading to poor results. We assume that this issue likely stems from the necessity for point MLLM to undergo training on point-text instruction data with trainable LLM parameters to enhance performance.

\begin{table*}
\centering
\begin{tabular}{l|l|cc|cc}
\toprule
\multirow{2}{*}{\textbf{Task}} & \multirow{2}{*}{\textbf{Method}} & \multicolumn{2}{c|}{\textbf{Instruction-type}} & \multicolumn{2}{c}{\textbf{Completion-type}} \\
& & P & P + I & P & P + I \\
\midrule
 & 3D-LLM & - & 49.00 & - & 41.50 \\
 & Point-LLM & 55.00 & - & 51.00 & - \\
 & ImageBind-LLM & * & 31.00 & * & 26.50 \\
\multicolumn{1}{l|}{\cellcolor{white} Objaverse} & X-InstructBLIP & * & 50.00 & * & 31.50 \\
\rowcolor{gray!10}
\multicolumn{1}{l|}{\cellcolor{white}} & Proj-only & 17.50* & 48.00 & 15.00* & 42.50 \\
\rowcolor{gray!10}
\multicolumn{1}{l|}{\cellcolor{white}} & \baseline & 56.00 & 55.00 & 54.50 & 59.50  \\
\rowcolor{gray!10}
\multicolumn{1}{l|}{\cellcolor{white}} & \ours & 57.00 & \textbf{60.50} & 56.50 & \textbf{62.00}  \\
\midrule
 & Point-LLM & 53.44 & - & 51.82 & - \\
 & ImageBind-LLM & * & 42.71 & * & 42.46 \\
 & X-InstructBLIP & * & 61.43 & * & 61.14 \\
\rowcolor{gray!10}
\multicolumn{1}{l|}{\cellcolor{white} ModelNet40} & Proj-only & 6.52* & 62.88 & 5.92* & 61.99 \\
\rowcolor{gray!10}
\multicolumn{1}{l|}{\cellcolor{white}} & \baseline & 68.59 & 66.00 & 58.79 & 64.59 \\
\rowcolor{gray!10}
\multicolumn{1}{l|}{\cellcolor{white}} & \ours & 68.76 & \textbf{70.02} & 58.71 & \textbf{65.24} \\
\bottomrule
\end{tabular}
\caption{Experimental results on zero-shot 3D object classification tasks using point (P) and combined point and image (P + I) inputs. Following \citet{xu2023pointllm}, two different types of prompts are considered: Instruction-type (``What is this?'') and Completion-type (``This is an object of''). *: models failed to follow open-ended generation instructions, leading to particularly low scores. }
\label{tab:full_point}
\end{table*}  

\section{Additional Qualitative Results}
\label{appendix:additional_qr}

\begin{figure*}[t]
    \centering
    \includegraphics[width=0.95\textwidth]{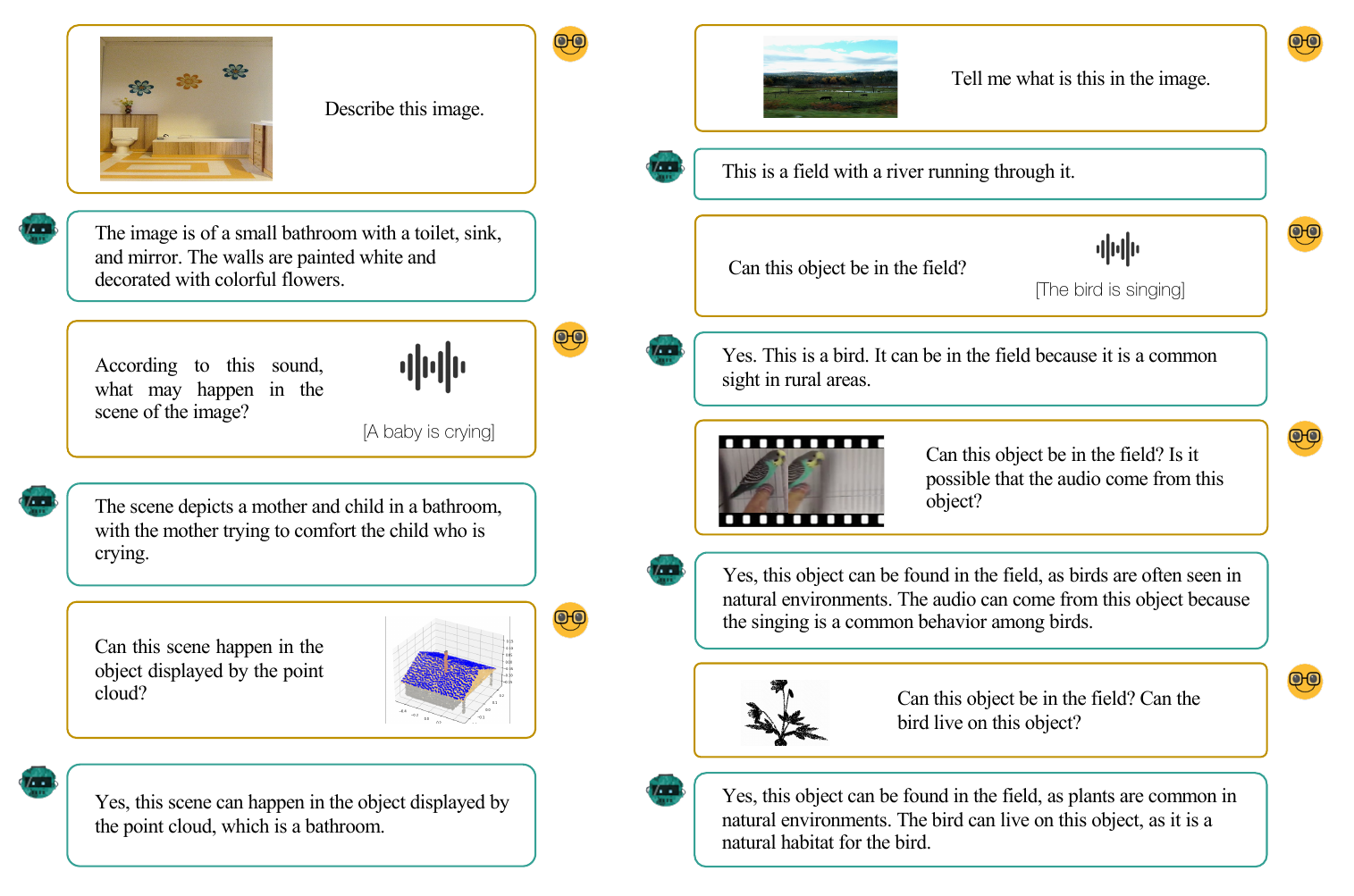}
    \caption{Additional qualitative results.}
    \label{fig:additional_qr1}
\end{figure*}

\begin{figure*}[t]
    \centering
    \includegraphics[width=0.95\textwidth]{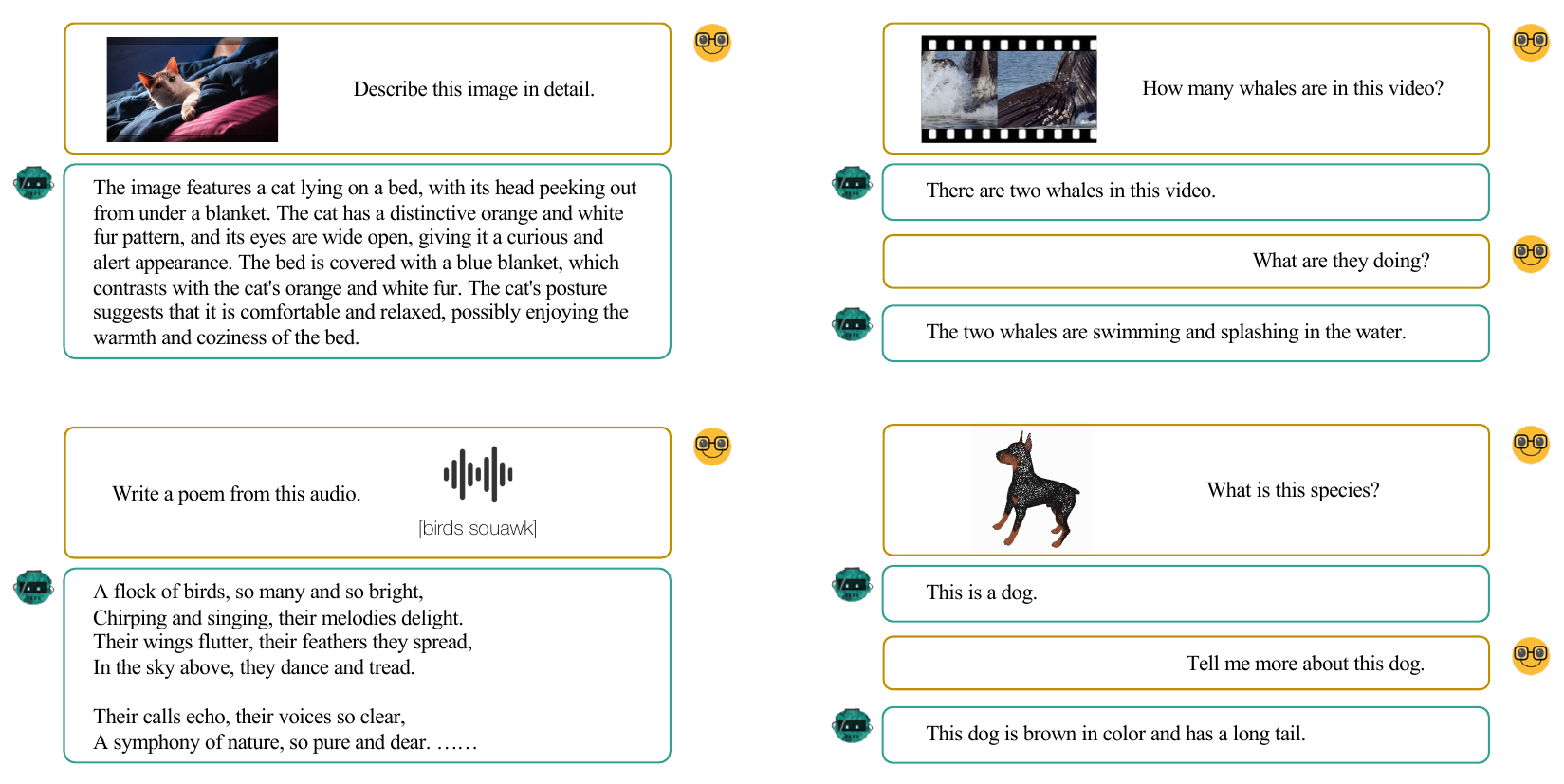}
    \caption{Additional qualitative results.}
    \label{fig:additional_qr2}
\end{figure*}

\begin{figure*}[t]
    \centering
    \includegraphics[width=0.95\textwidth]{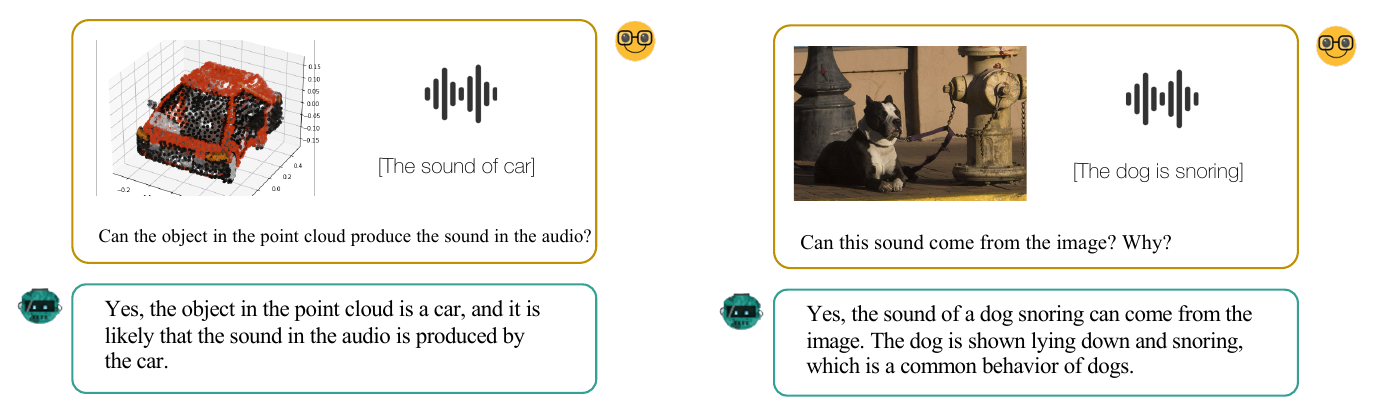}
    \caption{Additional qualitative results.}
    \label{fig:additional_qr3}
\end{figure*}

We provide additional qualitative results about in Figure \ref{fig:additional_qr1}-\ref{fig:additional_qr3}.

\end{document}